\documentclass[sigconf, nonacm]{acmart}

\newcommand\vldbpagestyle{plain} 
\usepackage[skins]{tcolorbox}
\usepackage{snaptodo}
\usepackage{amsmath, amssymb, amsthm}
\usepackage{graphicx}
\usepackage{hyperref}
\usepackage{enumitem}
\usepackage{mathrsfs}
\usepackage{xspace}

\usepackage{algorithm}
\usepackage{algorithmic}

\usepackage{tikz}
\usetikzlibrary{positioning, arrows.meta, shapes, calc}
\usepackage{dblfloatfix}

\newcommand{\model}{\textsc{auGraph}\xspace}

\begin{document}
\title{From Features to Structure: Task-Aware Graph Construction for Relational and Tabular Learning with GNNs}

\author{Tamara Cucumides}
\affiliation{%
  \institution{University of Antwerp}
  \country{Belgium}
}
\email{tamara.cucumidesfaundez@uantwerp.be}

\author{Floris Geerts}
\orcid{}
\affiliation{%
  \institution{University of Antwerp}
  \country{Belgium}
}
\email{floris.geerts@uantwerp.be}

\begin{abstract}
Tabular and relational data remain the most ubiquitous formats in real-world machine learning applications, spanning domains from finance to healthcare. Although both formats offer structured representations, they pose distinct challenges for modern deep learning methods, which typically assume flat, feature-aligned inputs. Graph Neural Networks (GNNs) have emerged as a promising solution by capturing structural dependencies within and between tables. However, existing GNN-based approaches often rely on rigid, schema-derived graphs---such as those based on primary-foreign key links---thereby underutilizing rich, predictive signals in non key attributes. In this work, we introduce \model, a unified framework for \emph{task-aware} graph augmentation that applies to both tabular and relational data. \model enhances base graph structures by selectively promoting attributes into nodes, guided by scoring functions that quantify their relevance to the downstream prediction task. This augmentation preserves the original data schema while injecting task-relevant structural signal. Empirically, \model outperforms schema-based and heuristic graph construction methods by producing graphs that better support learning for relational and tabular prediction tasks.

\end{abstract}

\maketitle

\pagestyle{\vldbpagestyle}

\section{Introduction}
Relational databases are a foundational data storage paradigm, widely used to manage structured data across industry and science. Their normalized schema—multiple interlinked tables connected via primary-foreign key relationships—supports modularity and efficient querying. 
However, this structure poses challenges for machine learning methods that assume flat, feature-complete inputs. The prevailing workaround, manual joins, feature engineering and aggregations, is not only laborious and memory intensive due to redundancy from denormalization, but also eliminates the relational structure that underpins the data, potentially leading to information loss and bias \citep{nargesian2017table,kipf2016semi}.


This has motivated recent research into automated, structure-aware methods for learning directly from relational without collapsing it into a single table \citep{cappuzzo2020embdi,tabgnn2021,pet2022,MullerPSNA19}. Even in the tabular setting, where data is already in a single table, graph-based approaches are used to model implicit structure \citep{li2024gnn4tabular}. These approaches construct graphs based on co-occurrence statistics \citep{cappuzzo2020embdi}, feature-wise multiplex connectivity \citep{tabgnn2021}, attribute-level hypergraphs \citep{pet2022}, or contextual linkage for specific tasks like conversational QA \citep{MullerPSNA19}. Yet, they generally lack task-aware mechanishms to guide graph construction.

Recent advances in \emph{Relational Deep Learning} (RDL) provide an alternative to traditional feature engineering. RDL frameworks transform relational databases into heterogeneous graphs—\emph{relational entity graphs}—where rows become nodes and key-based links define edges \citep{fey2023relational}. Graph Neural Networks are then applied to learn task-specific representations over these structures in an end-to-end fashion \citep{fey2023relational,schlichtkrull2018modeling,cvitkovic2019supervised}. This approach has achieved strong performance across multi-table prediction tasks, and has recently been extended with relational graph transformers \citep{dwivedi2025relational} and foundation models for in-context learning over arbitrary databases \citep{fey2025kumorfm}.

At the core of RDL and related GNN-based tabular learning lies a persistent challenge: how to construct a graph that captures meaningful statistical dependencies. Schema-derived graphs rely on key-based connectivity, often resulting in sparse or semantically limited structures. In single-table settings, various graph constructions exist, but they are typically task-agnostic and not optimized for downstream performance.

\textbf{Our contribution.} We present \model, a method that systematically integrates attribute-level signals into the graph structure through a task-guided augmentation process. Unlike approaches that rely solely on fixed schema links or indiscriminate feature inclusion, \model evaluates and incorporates attributes based on their utility for a given prediction task. This results in graph structures that are data-efficient, and tailored to the learning objective, applicable across both relational databases and single-tables.


\section{Preliminaries}
Here we first briefly recall the relational data model and then define the corresponding relational entity graphs, or REGs for short.

\begin{figure*}[t]
\centering
\includegraphics[width=\textwidth]{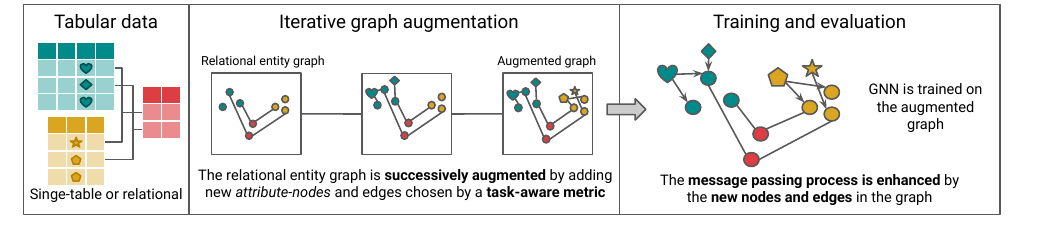}
\vspace*{-1cm}
\caption{
Overview of \model. Given a tabular dataset and a task, \model constructs a task-aware graph through iterative attribute promotion and scoring. The resulting graph is then used by a downstream GNN to perform the learning task.
}
\label{fig:fignn-overview}
\end{figure*}

\subsection{Databases} 
For $n\in\mathbb{N}$, we write $[n]=\{1,\dots,n\}$. 
Let $\mathsf{Dom}$ be an arbitrary \emph{domain} and $\mathcal{R}=\{R_1,\dots,R_n\}$ be a \emph{schema}, where each \emph{relation} $R_i$ has \emph{arity} $d_i\in\mathbb{N}$ and where we refer to $[d_i]:=\{1,2,\ldots,d_i\}$ as the \emph{attributes} of $R_i$. A \emph{table} $T_i\subseteq\mathsf{Dom}^{d_i}$ consists of \emph{tuples} (or \emph{rows}) $r=(r_1,\dots,r_{d_i})\in \mathsf{Dom}^{d_i}$ and we impose a \emph{primary-key constraint} on the first attribute: for all $r,s\in T_i$, $r_1=s_1$ implies $r=s$.
Let $\mathcal{L}\subseteq\{(R_i,j,R_k)\mid R_i,R_k\in\mathcal{R},\,j\in[d_i]\}$ be a set of \emph{foreign-key links}. On the table level, each $(R_i,j,R_k)\in\mathcal{L}$ enforces that for every $r\in T_i$ there exists $s\in T_k$ with $
r_j=s_1$, i.e.\ the $j$th attribute of $R_i$ references the primary key of $R_k$. We refer to $(\mathcal{R},\mathcal{L})$ as a \emph{relational schema} and to the corresponding tables $D=(T_1,\ldots,T_n)$ satisfying the key and foreign-key constraints as a \emph{database} of that schema.

\smallskip
$\rhd$ \emph{Remark.} We observe that \emph{tabular data} can be regarded as a special instance of a database, consisting of a single table $T$ of schema $\mathcal R=\{R\}$ and with no foreign-key links ($\mathcal L=\emptyset)$. 
\smallskip


\subsection{Relational Entity Graphs}
Given a schema $(\mathcal{R},\mathcal{L})$ and a database $D$ over that schema, we define its \emph{relational entity graph} as
$G_{\text{REG}}= (V,E,\phi,\tau)$, where
\begin{itemize}[noitemsep]
\item the vertex set $V$ is defined as
$
V:= \{\,v_r \mid r\in T_i,\;R_i\in\mathcal{R}\}
$, i.e., $V$ contains one vertex $v_r$ for each tuple $r$ in each table in $D$;
\item the edge set $E$ is given by
$
E:= \{(v_r,v_s)\mid (R_i,j,R_k)\in\mathcal{L},\;r\in T_i,\;s\in T_k,\;r_j=s_1\}
$, 
i.e., $E$ contains an edge $(v_r,v_s)$ whenever the corresponding tuples satisfy a foreign-key link $(R_i,j,R_k)\in\mathcal{L}$; and  
\item $\phi$ and $\tau$ label each vertex $v_r$ by its tuple $r\in T_i$ and relation name $R_i$ of $T_i$, respectively, i.e.,
$\phi(v_r):=r$ and $\tau(v_r):=R_i$.
\end{itemize}

\smallskip
$\rhd$ \emph{Remark.} It should be clear that we can move freely between schemas and databases, and relational entity graphs. That is, they carry the same information. Of course, the connection between tuples is explicit in the graph representation, whereas it is implicit in the database representation.

\subsection{Prediction Task}
We focus on \emph{tuple‐level} (or node‐level) \emph{prediction tasks}. Given a relation $R_i\in\mathcal R$ with corresponding table $T_i$ in database $D$, we aim to learn a function $f\colon T_i\to\mathcal Y$ for some output space $\mathcal Y$. To encode this relationally, we extend the relational schema by adding a relation $R_{\mathrm{train}}$ of arity 2 whose table 
$T_{\mathrm{train}}$ consists of pairs $(r_1,f(r))$ of tuple keys $r_1$ of tuples $r\in T_i$ and their labels $f(r)\in \mathcal Y$. We also add the foreign‐key link $(R_{\mathrm{train}},1,R_i)\in\mathcal L$, enforcing that the first attribute of $R_{\mathrm{train}}$ references the primary key of $R_i$. In the relational entity graph $G_{\text{REG}}=(V,E,\phi,\tau)$, this creates a set of training vertices $V_{\mathrm{train}}=\{v_r\mid r\in T_i\}$, each labeled by its key and true label. For simplicity, we use $f(v)$ to denote its true label 
encoded by $\phi(v)_2$.

We implement or learn $f$ using graph neural networks, which we abstract as functions $f_\theta(G,v)\in\mathcal Y$ taking a graph $G$ and a vertex $v$ as input and returning a value in $\mathcal Y$, with $\theta$ (weights, biases, etc.) in some parameter space $\Theta$. Training amounts to finding $\theta\in\Theta$ that minimize the empirical loss $\sum_{v\in V_{\mathrm{train}}}\ell\bigl(f_\theta(G,v),f(v)\bigr)$ on the graph $G$ and training vertices for some loss function $\ell$.


\section{Feature informed graph construction}

We introduce \model, a principled and scalable framework for \emph{task-aware graph construction} from relational and tabular data. Our key insight is to treat graph construction itself as a pre-processing step tuned for learning—augmenting the base entity graph by promoting high-value attributes into new structural components. These attribute-derived nodes and edges are selected via scoring functions that capture statistical, structural, and model-based signals of predictive relevance. By iteratively augmenting the graph based on these signals, \model tailors the graph topology to better support the target task, all while avoiding brute-force enumeration of attribute subsets. As shown in Figure~\ref{fig:fignn-overview}, this graph augmentation process enhances message passing and improves GNN performance, producing graphs that are compact, interpretable, and aligned with the learning objective.
We now describe the main components of \model: how attribute promotion augments the graph structure, the scoring functions used to evaluate candidate attributes, and how these are combined in the iterative construction procedure.

\subsection{Augmenting Graphs with Attribute Nodes}
In terms of databases, relational entity graph augmentation corresponds to extracting a new unary table of constants from a chosen attribute and linking it back to the original graph. More precisely, let $(\mathcal{R},\mathcal{L})$ be a relational schema with database $D$. Fix $R_i\in\mathcal{R}$ of arity $d_i$ and an attribute $A\in[d_i]$. We introduce a new unary relation $R_A$ with table
$
T_A:=\pi_A(T_i)=\{\,r_A\mid r\in T_i\}
$,
and add the foreign‐key link $(R_A,1,R_i)$ to $\mathcal{L}$. 
In graph terms, starting from the relational entity graph $G_{\mathrm{REG}}=(V,E,\phi,\tau)$ of $(\mathcal{R},\mathcal{L})$ and $D$, the augmented graph $G_{\mathrm{REG}}^{(A)}$ is obtained by
\begin{itemize}[noitemsep]
  \item adding vertices $V_A = \lbrace v_a : a\in T_A \rbrace$, i.e., one vertex per each attribute value;
  \item for each original vertex $v\in V$ with $\phi(v)=r$, $\tau(v)=R_i$ and $r_A=a$, adding an edge $(v_a,v)$; and finally,
  \item setting $\phi(v_a):=a$ and $\tau(v_a):=R_A$.
\end{itemize}
If multiple attributes $A_1,A_2,\dots$ are promoted, the resulting graph is denoted $G_{\mathrm{REG}}^{(A_1,A_2,\dots)}$ and is obtained by sequential augmentation.

\subsection{Attribute Scoring Metrics} \label{sec:scoring-metrics}

To guide graph augmentation, \model uses task-aware scoring functions that rank attributes by their potential to improve downstream prediction. Each score captures a different signal: from raw statistical relevance to structural and model-informed criteria. We recall that $V_{\text{train}}$ is the set of vertices in our graph, corresponding to the tuples in some relation for which we have the actual labels.


Given a non-key attribute $A$, we evaluate its relevance for the learning task via the following metrics: 

\medskip
\noindent\textbf{Statistical Signal.} We begin with a classical feature selection metric, which does not take into account the graph structure:

\noindent
$\rhd$ \emph{Mutual Information.} We define the \emph{fully joined training table} of  \( D \) as \( \mathbf{T}_{\text{train}} = T_1 \bowtie \cdots \bowtie T_n \bowtie T_{\text{train}} \), where the joins are computed based on the foreign-key links, and compute the mutual information between the attribute \( A \) and the label \( \mathcal{Y} \):
\[
s_{\text{MI}}(A) = \sum_{(a,y) \in \pi_{A,Y}(\mathbf{T}_{\text{train}})}  \hat{P}(a, y) \log \left( \frac{\hat{P}(a, y)}{\hat{P}(a) \hat{P}(y)} \right),
\]
where \( \hat{P}(a, y) \) denotes the empirical joint probability of observing value \( a \) for attribute \( A \) and label \( y \) in \( \mathbf{T}_{\text{train}} \), and the empirical marginals are given by \( \hat{P}(a) = \sum_y \hat{P}(a, y) \) and \( \hat{P}(y) = \sum_a \hat{P}(a, y) \). This score quantifies the dependence between the attribute and the label, and is computed directly from the tabular data.

\medskip
\noindent\textbf{Graph-aware Signal.}
We next assess whether $A$ improves the topology of the graph for learning.

\noindent
$\rhd$ \emph{Entropy Gain.} The entropy gain metric measures how graph augmentation with attribute $A$ creates more homogeneous neighborhoods with respect to the target labels in  $\mathcal{Y}$

For a node \( v \), let \( \mathsf{N}_G^d(v) \) denote its \( d \)-hop neighborhood in graph \( G \), and let \( \mathsf{N}_{G,\text{train}}^d(v) = \mathsf{N}_G^d(v) \cap V_{\text{train}} \) be the subset of training nodes within it. We define the empirical label distribution over a set $S \subseteq V_{\text{train}}$ as $\hat{P}_S(y) = \frac{1}{|S|} \sum_{u \in S} \mathbb{I}[f(u) = y]$, and its entropy:
\[
H(S) = -\sum_{y} \hat{P}_S(y) \log \hat{P}_S(y).
\]
Then, the entropy gain score can be written as:
\[
s_{\text{ent}}(A) = \frac{1}{|V_{\text{train}}|} \sum_{v \in V_{\text{train}}} \left[ H\big(\mathsf{N}^{d}_{G_{\text{REG}},\text{train}}(v) \big) - H\big(\mathsf{N}^{d}_{G_{\text{REG}}^{(A)},\text{train}}(v)\big) \right].
\]

\noindent
$\rhd$ \emph{Path Disagreement.} Let $\mathsf{Path}_A(u, v) = 1$ if $u$ and $v$ in $V_{\text{train}}$ are connected through a shared attribute-vertex in $V_A$, and $0$ otherwise. Then we define the path disagreement metric as:
\[
s_{\text{dis}}(A) = \frac{1}{Z} \sum_{\substack{u, v \in V_{\text{train}}}} \mathsf{Path}_A(u, v) \cdot \mathbb{I}[f(u) \ne f(v)],
\]
where $Z$ is a normalization constant. Lower values suggest that $A$ tends to connect nodes with the same label.

\medskip
\noindent\textbf{Model-based Signal.}
We also include a GNN-based score.

$\rhd$ \emph{GNN Gain.} A model \( f_\theta \), which maps nodes in a graph to predictions, is trained once on \( G_{\text{REG}} \) and then kept fixed. We assess how augmenting the graph with attribute \( A \) affects its performance over the validation nodes:
\[
s_{\text{GNN}}(A) = \mathsf{Eval}_{f_\theta}(G_{\text{REG}}^{(A)}) - \mathsf{Eval}_{f_\theta}(G_{\text{REG}}),
\]
where \( \mathsf{Eval}_{f_\theta}(G) \) denotes the aggregated validation performance of \( f_\theta \) on graph \( G \), using a task-specific metric (e.g., accuracy or F1 score). This provides a lightweight proxy for the utility of attribute \( A \): if its inclusion improves message passing, performance should increase—even without retraining.

\medskip
While some scores are purely statistical, and others are structure or model-aware, all are designed to promote attributes that support downstream learning.
\subsection{Graph Augmentation Procedure}
Rather than selecting a fixed top-$k$ set of attributes in one pass, \model builds the graph incrementally through a selection loop. Starting from the base graph $G_{\text{REG}}$, it repeatedly selects the highest scoring attribute (according to a chosen scoring function) and promotes it into the graph. After each promotion, scores are recomputed to reflect the updated structure, favoring attributes that are complementary rather than redundant. This process continues until a fixed budget $k$ is reached or the best available score falls below an ealrly-stopping threshold $\tau$. The resulting graph $G_{\text{REG}}^{(A_1, \dots, A_t)}$  can then be used as input to any downstream GNN model.

\section{Experiments}
We test \model in three settings, focusing on:
\begin{enumerate}[leftmargin=*,label=(Q\arabic*),topsep=1pt,itemsep=1pt]
    \item \textbf{Attribute discovery}: Can \model identify task-relevant attributes, and does promoting them improve performance?
    \item \textbf{Relational augmentation}: Can \model enhance schema-derived graphs in multi-table relational databases?
    \item \textbf{Tabular graph quality}: Does \model outperform heuristic graph construction baselines on single-table data?
\end{enumerate}
The implementation of \model is available in \url{https://github.com/TamaraCucumides/auGraph}
\subsection{Setup}
We evaluate node classification performance across three regimes: (i) synthetic relational data, (ii) single-table tabular benchmarks, and (iii) multi-table relational databases. In each case, data is transformed into a graph via a designated construction strategy, and a fixed GNN model is trained for supervised learning. Only the graph structure varies between runs.

The GNN is a two-layer heterogeneous architecture implemented with \texttt{HeteroConv} and \texttt{SAGEConv} from PyTorch Geometric~\cite{fey2019pyg}, using mean aggregation~\cite{hamilton2017graphsage}. Training hyperparameters and loss are held constant across all experiments. 

We compare \model by training and evaluating a GNN over the following graph constructions:
\begin{itemize}[leftmargin=*,topsep=0pt,itemsep=1pt]
    \item \textbf{REG}\footnote{The \textbf{REG} baseline is not included for tabular data, as it would result in a set of disconnected nodes. }: Relational Entity Graph.~\cite{fey2023relational}
    \item \textbf{All-promote}: Graphs in which all attributes are promoted~\cite{zhang2023gfs}.
    \item \textbf{Random-$k$}: Graphs built by promoting $k$ random attributes.
    \item \textbf{kNN Graph}: Instance similarity graphs constructed via $k$-nearest neighbors in feature space~\cite{ccns2023}. 
\end{itemize}

We evaluate models using accuracy, F1-score, and ROC-AUC. Accuracy and F1 reflect classification quality; ROC-AUC captures ranking performance and robustness to class imbalance.

\vspace{0.5em}
\noindent\textbf{Datasets.} We use three different data regimes:
\begin{itemize}[leftmargin=*,topsep=0pt,itemsep=1pt]
    \item \textbf{Synthetic:} A relational schema modeled after \texttt{relbench-h\&m}~\cite{fey2023relational}, with a binary label defined by a known subset of attributes. This tests \model ability to recover relevant structure (Q1) and improve over relational entity graphs (Q2).

    \item \textbf{Relational:} The \texttt{hepatitis} dataset from the CTU Relational Datasets~\cite{motl2015ctu}, where the task is to classify entities in a multi-table schema. We compare \model against relational entity graph, all-promote, and random baselines (Q2).

    \item \textbf{Tabular:} The UCI \texttt{mushroom} dataset~\cite{mushroom_73}, treated as a flat table without schema. This setting evaluates whether \model can construct effective graph structure from purely tabular data (Q3).
\end{itemize}

\subsection{Results and Discussion}
\begin{figure}[ht]
\centering
\includegraphics[width=0.47\textwidth]{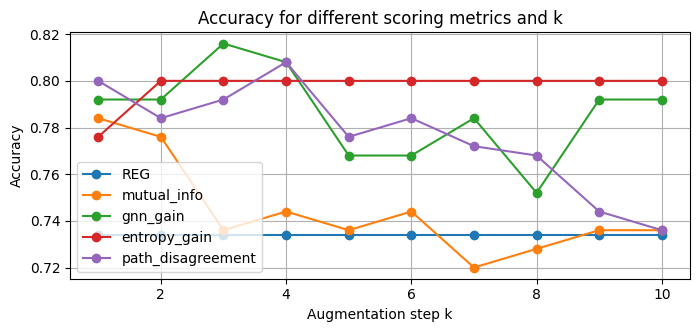}
    \caption{Accuracy on augmented graph with different scoring methods versus promotion budget $k$. Synthetic data.}
    \label{fig:scoring-k-synthetic}
\end{figure}

To evaluate (Q1), we focus on synthetic data, where the ground-truth label depends on three specific attributes. We perform graph augmentation using all scoring metrics and examine which attributes each method promotes. With a promotion budget of $k=3$, the \texttt{path-disagreement} metric identifies two of the three ground-truth attributes but ranks third in downstream performance, suggesting that recovering relevant features does not necessarily yield the most effective graph structure for learning. The \texttt{gnn-gain} metric, while promoting only one relevant attribute, achieves the highest performance by aligning directly with message passing (see Figure~\ref{fig:scoring-k-synthetic}). The \texttt{entropy-gain} score plateaus after two promotions and stops, never degrading performance—suggesting greater robustness. Notably, all metrics outperform the minimal relational entity graph, showing their general utility for guiding augmentation.

\begin{table}[t]
\centering
\caption{Performance in test data. Results are reported for $k=3$ and the Random-$k$ is averaged over 3 runs. \model results are from the metric that performs best on the validation set.}
\label{tab:results}
\begin{tabular}{lccc}
\toprule
\textbf{Model} & \textbf{F1-score} & \textbf{Accuracy} & \textbf{ROC-AUC} \\
\midrule
\multicolumn{4}{c}{\textit{Synthetic Data (Relational)}} \\ \midrule
REG         & 0.734 & 0.784 & 0.834 \\
All-promote      & 0.668 & 0.728 & 0.726 \\
Random-$k$       & 0.699 & 0.765 & 0.786 \\
\model (top-$k$, $s_{\text{GNN}}$)       & \bf 0.773 & \bf 0.816 & \bf 0.836 \\
\midrule
\multicolumn{4}{c}{\textit{Relational Data}} \\ \midrule
REG          & 0.930 & 0.936 & 0.987 \\
All-promote      & 0.894 & 0.904 & 0.985 \\
Random-$k$       & 0.917 & 0.924 & \bf{0.988} \\
\model (top-$k$, $s_{\text{GNN}}$)    & \textbf{0.939} & \textbf{0.944} & 0.987 \\ \midrule
\multicolumn{4}{c}{\textit{Tabular Data}} \\ \midrule
All-promote      & 0.969 & 0.970 & \bf 0.992 \\
Random-$k$       & 0.944 & 0.946 & 0.987 \\
kNN Graph (k=10)       & 0.937 & 0.936 & 0.979 \\
\model (top-$k$, $s_{\text{MI}}$)    & \textbf{0.985} & \textbf{0.985} & \textbf{0.992} \\

\bottomrule
\end{tabular}
\end{table}

To address (Q2) and (Q3), we refer to Table~\ref{tab:results}, which compares performance across relational and tabular benchmarks.

In the relational setting, across both synthetic and real datasets, \model consistently outperforms the baseline relational entity graph. Naively promoting all attributes degrades performance, often underperforming even the minimal graph, while random promotion yields modest gains. These results show that structural complexity alone is insufficient; task-aware augmentation is crucial. \model's consistent improvements demonstrate that its scoring functions successfully identify attributes that induce useful inductive bias.

In the tabular setting, where the base graph is often empty or poorly structured, promoting random attributes proves ineffective, likely due to arbitrary edge formation. While promoting all attributes helps, it also brings redundancy and noise. \model again achieves the best performance, showing that feature-informed selective augmentation enables compact yet expressive graphs that improve downstream learning.

Finally, we emphasize the importance of the promotion budget $k$ and early stopping threshold $\tau$. Promoting low-quality attributes can obscure the signal, but \model's controlled, iterative augmentation mitigates this risk. Both $k$ and $\tau$ are selected via validation performance, ensuring efficient and robust graph construction.
\section{Conclusions and future directions}

\model introduces a principled, task-aware alternative to schema-based or heuristic graph construction for tabular and relational data. By promoting informative attributes into the graph topology, \model consistently improves downstream performance. To our knowledge, this is the first framework to pose graph construction as a task-aware feature selection problem.

Looking ahead, an important direction is to extend \model to \emph{temporal relational data}. While current scores apply to temporally-aware graphs, they fail to capture unique dynamics—such as evolving attributes, cross-time entity links, or timestamped dependencies. Designing temporal-sensitive scores could yield stronger structural priors and more expressive graphs.

A second and more foundational direction is to better understand the \emph{expressivity and generalization} of task-aware graph construction. A compelling goal is to align augmentation more directly with labels—e.g., by maximizing agreement between 1-WL colourings and targets. This could define a notion of \emph{task-optimal graph construction}, offering both a theoretical target and a practical guide. Understanding how structural properties shape generalization remains an open challenge, especially given the weak empirical link between expressivity and performance in GNNs~\cite{morris2023generalization}. We view this as a promising and underexplored research avenue.

\newpage
\bibliographystyle{ACM-Reference-Format}
\bibliography{references}
\end{document}